\documentclass{article}
\usepackage{spconf,amsmath,graphicx}
\usepackage{amssymb,amsfonts}
\usepackage{algorithmic}
\usepackage{soul}
\usepackage{marvosym}
\usepackage{mathtools}
\usepackage{multicol}
\usepackage{colortbl}
\usepackage[]{xcolor}
\usepackage{float}
\usepackage{subfigure}
\usepackage{graphicx}
\usepackage{caption}
\usepackage{color}
\usepackage{svg}
\usepackage{pifont}
\usepackage{booktabs}
\usepackage{makecell}
\usepackage{multirow}
\usepackage{algorithm}
\usepackage{color}
\usepackage{comment}

\title{DMKD: Improving Feature-based Knowledge Distillation for Object Detection Via Dual Masking Augmentation}

\name{Guang Yang$^{1\dagger}$, Yin Tang$^{2\dagger}$\thanks{$^{\dagger}$ Equal Contribution}, Zhijian Wu$^{3}$, Jun Li$^{1}$\textsuperscript{(\Letter)}\thanks{\textsuperscript{\Letter} Corresponding Author}, Jianhua Xu$^{1}$, Xili Wan$^{2}$}

\address
{
{$^1$ School of Computer and Electronic Information, Nanjing Normal University, Nanjing, China
}\\
{$^2$ College of Computer and Information Engineering, Nanjing Tech University, Nanjing, China
}\\
{$^3$ School of Data Science and Engineering, East China Normal University, Shanghai, China
}
}

\begin{document}
%

%

\maketitle
\begin{abstract}
Recent mainstream masked distillation methods function by reconstructing selectively masked areas of a student network from the feature map of its teacher counterpart. In these methods, the masked regions need to be properly selected, such that reconstructed features encode sufficient discrimination and representation capability like the teacher feature. However, previous masked distillation methods only focus on spatial masking, making the resulting masked areas biased towards spatial importance without encoding informative channel clues. In this study, we devise a Dual Masked Knowledge Distillation (DMKD) framework which can capture both spatially important and channel-wise informative clues for comprehensive masked feature reconstruction. More specifically, we employ dual attention mechanism for guiding the respective masking branches, leading to reconstructed feature encoding dual significance. Furthermore, fusing the reconstructed features is achieved by self-adjustable weighting strategy for effective feature distillation. Our experiments on object detection task demonstrate that the student networks achieve performance gains of 4.1\% and 4.3\% with the help of our method when RetinaNet and Cascade Mask R-CNN are respectively used as the teacher networks, while outperforming the other state-of-the-art distillation methods.
\end{abstract}

\section{INTRODUCTION}\label{sec:intro}
It is well known that knowledge distillation is capable of transferring the dark knowledge learned in a complex teacher network to a lightweight student network. The earlier knowledge distillation algorithms primarily focus on the output head of the networks, and the representative schemes include logit-based distillation for classification~\cite{hinton2015distilling, zhou2020rethinking} and head-based distillation for detection~\cite{chen2017learning,dai2021general,zhixing2021distilling}. More recently, feature-based distillation strategies have received increasing attention due to its desirable task-agnostic flexibility~\cite{heo2019comprehensive,chen2021distilling,chen2021cross,wang2020exclusivity}. In particular, feature distillation helps the student network imitate the teacher feature with enhanced representational capability. In object detection, a variety of advanced feature distillation methods have been developed. To be specific, the classic FitNet~\cite{romero2014fitnets} performed distillation at the global feature level. FGD~\cite{yang2022focal} was developed to distinguish foreground and background distillation within a unified framework. CWD~\cite{shu2021channel} simply minimized the Kullback-Leibler divergence~\cite{kullback1951information} between the channel probability maps of the two networks, and the distillation process is inclined to the most significant regions of each channel.

\begin{figure}[tb]
\centerline{\includegraphics[width=0.5\textwidth]{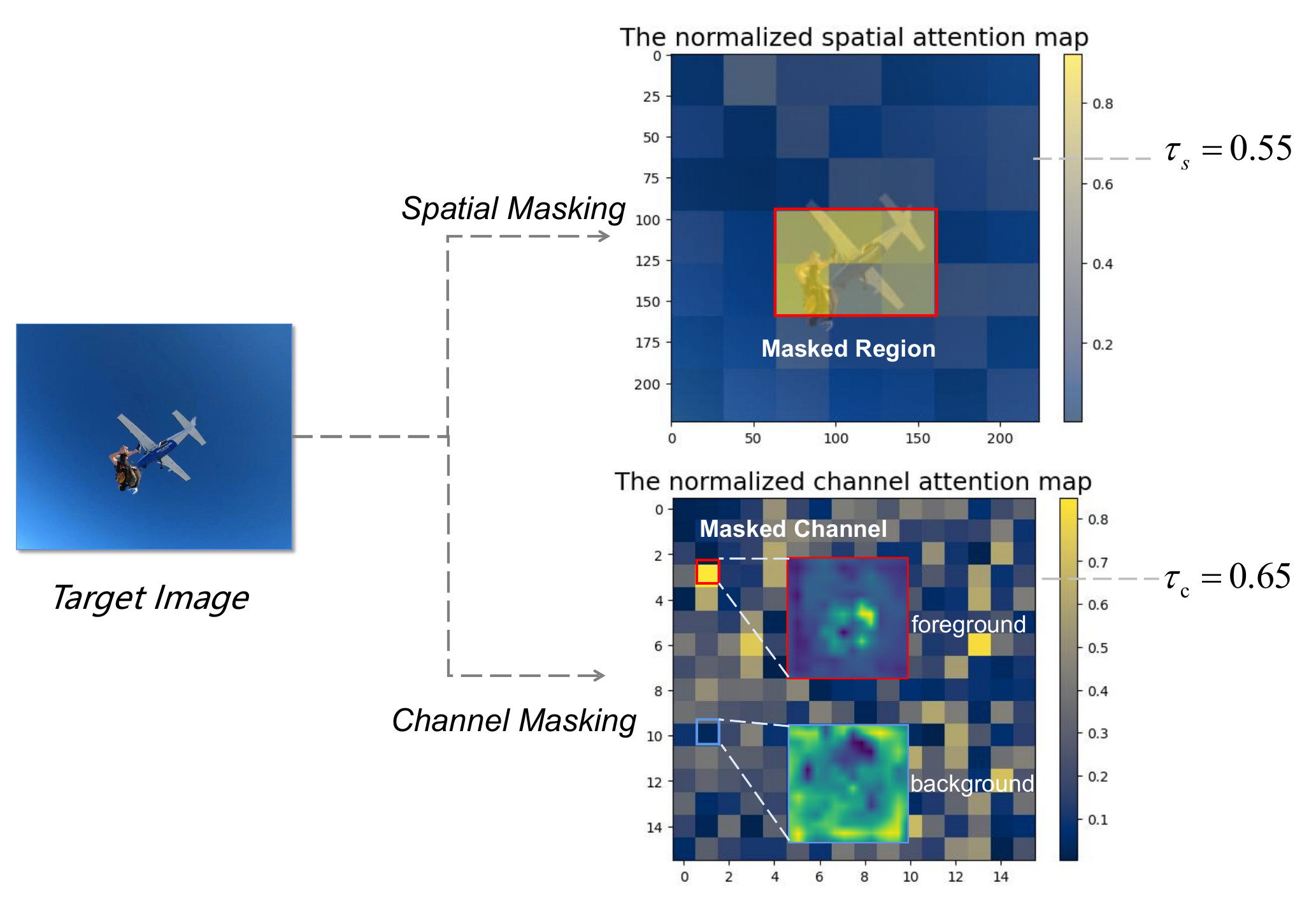}}
\caption{\textbf{Illustration of our dual masking strategies}. Obviously, our method can simultaneously capture the spatially salient and object-specific channel-wise clues which are highlighted on the spatial and channel attention map for dual masking. $\tau_{s}$ and $\tau_{c}$ are introduced as two hyperparameters which are used for adjusting the respective masking ratio. }
\label{mask_vis}
\end{figure}

In feature distillation methods, recent research suggests it is preferable for the student model to reconstruct important features from the teacher model in the first place instead of following the teacher for generating competitive representations. For instance, MGD \cite{yang2022masked} was proposed to randomly mask pixels in the feature map of the student network, leading to reconstructed features of the teacher model via a simple block. AMD \cite{10191080} performed attention-guided feature masking on the feature map of the student network, such that region-specific important features can be identified via spatially adaptive feature masking instead of random masking in the previous method.


\begin{figure}[t]
\centerline{\includegraphics[width=0.45\textwidth]{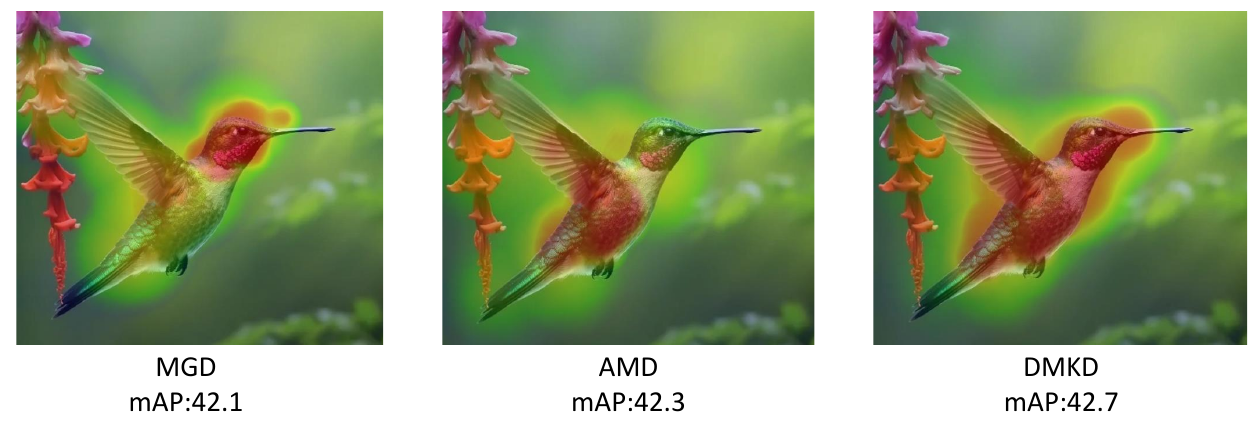}}
\caption{Visualisation of the feature maps obtained by different distillation methods. }
\label{vis}
\end{figure}

Although current masked feature distillation methods contribute to improving the performance of the student model, the selectively masked areas in the aforementioned approaches are generated only considering spatial importance. For instance, MGD~\cite{yang2022masked} performs random feature masking that is position-agnostic, resulting in non-informative and semantically irrelevant regions to be reconstructed. This limitation is somewhat alleviated in AMD~\cite{10191080} which utilizes spatial-wise masking for identifying the spatially salient regions, whereas it still fails to identify the informative channels which are critical for dense prediction task as indicated in \cite{shu2021channel, guo2022segnext}. In this study, differently, we attempt to capture spatially important regions and object-aware channel-wise clues both of which can be obtained by dual attention mechanisms. In particular, the dual masking strategy illustrated in Fig.~\ref{mask_vis} is used for masked feature reconstruction to comprehensively characterize the significant object-aware features. As shown in Fig.~\ref{vis}, in contrast to MGD and AMD which only capture discriminative local features (head and body of a bird), our DMKD provides a more comprehensive object-specific encoding for improved representation capability. 

\section{Methodology}\label{sec:method}

\begin{figure*}[tb]
\centerline{\includegraphics[width=1.0\textwidth]{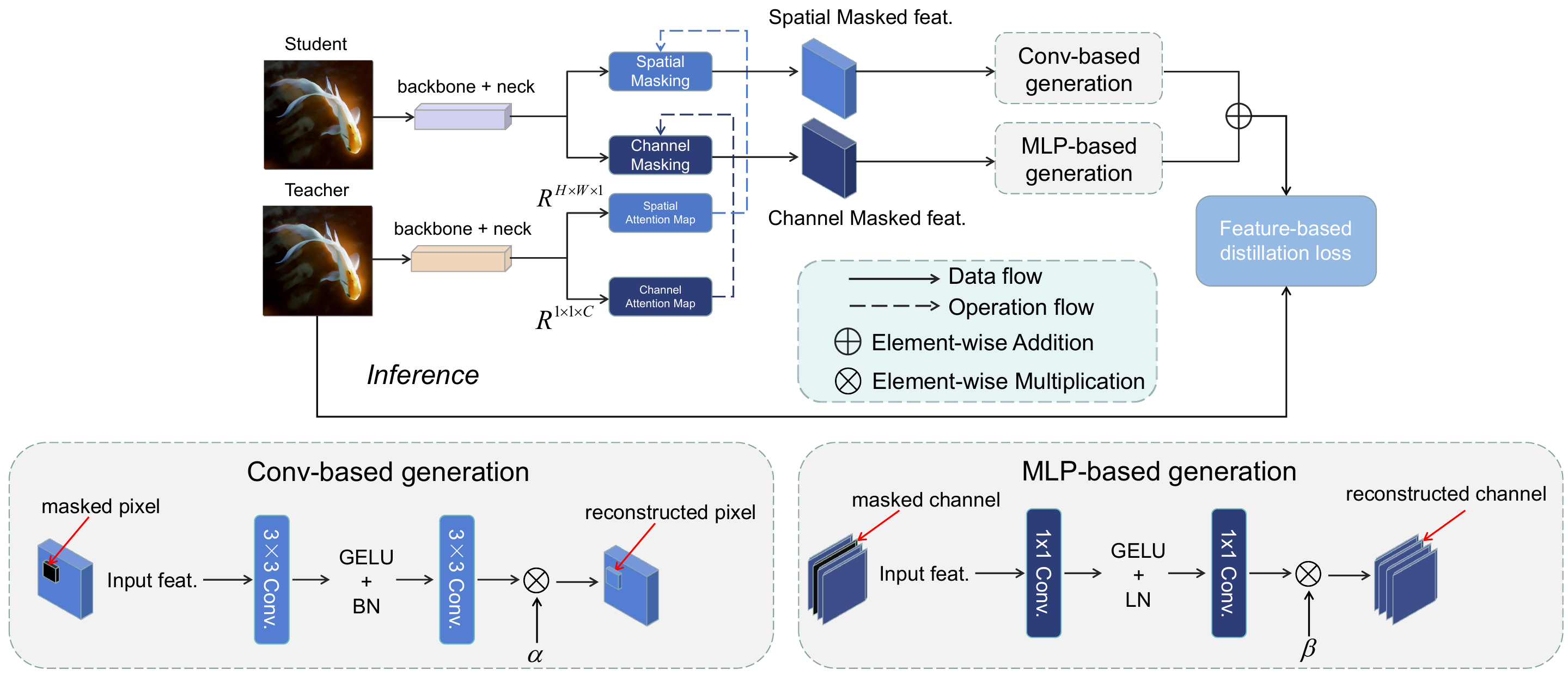}}
\caption{\textbf{Overall pipeline of the proposed DMKD.} After generating the spatial and channel-wise attention maps from the teacher feature, we perform dual masking guided by the attention maps for reconstructing important object-aware features. Finally, the reconstructed features are fused with two learnable parameters $\alpha$ and $\beta$ for self-adjustable weighting.}
\label{dmkd_arch}
\end{figure*}

\begin{table*}[thb]
\centering
\renewcommand{\arraystretch}{1.1}
\caption{Comparison of our method with other distillation methods for object detection on COCO.}
\resizebox{0.8\textwidth}{!}{%
\begin{tabular}{c|l|llll|llll}
\hline
\multicolumn{1}{c|}{Teacher}                                                                         & Student                      & mAP                                & $AP_{S}$                          & $AP_{M}$                          & $AP_{L}$                          & mAR                                & $AR_{S}$                          & $AR_{M}$                             & $AR_{L}$                              \\ \hline
                                                                                                     & RetinaNet-ResNet50 (\textit{\textcolor{gray}{baseline}})              & 37.4                               & 20.6                         & 40.7                         & 49.7                         & 53.9                               & 33.1                         & 57.7                         & 70.2                         \\
                                                                                                     & FKD \cite{zhang2020improve}                            & 39.6 (+2.2)                         & 22.7                         & 43.4                         & 52.5                         & 56.1 (+2.2)                         & 36.8                         & 60.0                         & 72.1                         \\
                                                                                                     & FGD \cite{yang2022focal}                           & 40.7 (+3.3)                         & 22.9                         & 45.0                         & 54.7                         & 56.8 (+2.9)                         & 36.5                         & 61.4                         & 72.8                         \\
                                                                                                     & MGD \cite{yang2022masked}                         & 41.0 (+3.6)                         & 23.4                         & 45.3                         & 55.7                         & 57.0 (+3.1)                         & 37.2                         & 61.7                         & 72.8                         \\
                                                                                                     & AMD \cite{10191080}                              & 41.3 (+3.9)                         & 23.9                         & 45.4                         & 55.7                         & 57.4 (+3.5)                         & 38.2                         & 61.7                         & 73.5                         \\
\multirow{-6}{*}{\begin{tabular}[c]{@{}c@{}}RetinaNet~\cite{lin2017focal}\\ ResNeXt101\\  (41.0)\end{tabular}}           & \cellcolor[HTML]{EFEFEF}\textbf{DMKD} (Ours) & \cellcolor[HTML]{EFEFEF}\textbf{41.5} (\textcolor{blue}{+4.1}) & \cellcolor[HTML]{EFEFEF}24.0 & \cellcolor[HTML]{EFEFEF}45.8 & \cellcolor[HTML]{EFEFEF}55.8 & \cellcolor[HTML]{EFEFEF}\textbf{57.8} (\textcolor{blue}{+3.9}) & \cellcolor[HTML]{EFEFEF}38.5 & \cellcolor[HTML]{EFEFEF}62.3 & \cellcolor[HTML]{EFEFEF}73.2 \\ \hline
                                                                                                     & RetinaNet-ResNet50 (\textit{\textcolor{gray}{baseline}})              & 38.6                               & 22.5                         & 42.2                         & 50.4                         & 55.1                               & 34.9                         & 59.4                         & 70.3                         \\
                                                                                                     & FKD \cite{zhang2020improve}                            & 40.6 (+2.0)                         & 23.4                         & 44.6                         & 53.0                         & 56.9 (+1.8)                         & 37.3                         & 60.9                         & 71.4                         \\
                                                                                                     & FGD \cite{yang2022focal}                           & 42.0 (+3.4)                         & 24.0                         & 45.7                         & 55.6                         & 58.2 (+3.1)                         & 37.8                         & 62.2                         & 73.3                         \\
                                                                                                     & MGD \cite{yang2022masked}                         & 42.3 (+3.7)                         & 24.4                         & 46.2                         & 55.9                         & 58.4 (+3.3)                         & 40/4                         & 62.3                         & 73.9                         \\
                                                                                                     & AMD \cite{10191080}                              & 42.7 (+4.1)                         & 24.8                         & 46.5                         & 56.3                         & 58.8 (+3.7)                         & 40.6                         & 62.4                         & 74.1                         \\
\multirow{-6}{*}{\begin{tabular}[c]{@{}c@{}}RepPoints~\cite{yang2019reppoints}\\ ResNeXt101\\ (44.2)\end{tabular}}            & \cellcolor[HTML]{EFEFEF}\textbf{DMKD} (Ours) & \cellcolor[HTML]{EFEFEF}\textbf{42.9} (\textcolor{blue}{+4.3}) & \cellcolor[HTML]{EFEFEF}25.1 & \cellcolor[HTML]{EFEFEF}46.9 & \cellcolor[HTML]{EFEFEF}56.4 & \cellcolor[HTML]{EFEFEF}\textbf{60.1} (\textcolor{blue}{+4.0}) & \cellcolor[HTML]{EFEFEF}40.9 & \cellcolor[HTML]{EFEFEF}62.9 & \cellcolor[HTML]{EFEFEF}74.4 \\ \hline
                                                                                                     & RetinaNet-ResNet50 (\textit{\textcolor{gray}{baseline}})              & 38.4                               & 21.5                         & 42.1                         & 50.3                         & 52.0                               & 32.6                         & 55.8                         & 66.1                         \\
                                                                                                     & FKD \cite{zhang2020improve}                            & 41.5 (+3.1)                         & 23.5                         & 45.0                         & 55.3                         & 54.4 (+2.4)                         & 34.0                         & 58.2                         & 69.9                         \\
                                                                                                     & FGD \cite{yang2022focal}                           & 42.0 (+3.6)                         & 23.8                         & 46.4                         & 55.5                         & 55.4 (+3.4)                         & 35.5                         & 60.0                         & 70.0                         \\
                                                                                                     & MGD \cite{yang2022masked}                          & 42.1 (+3.7)                         & 23.7                         & 46.4                         & 56.1                         & 55.5 (+3.5)                         & 35.4                         & 60.0                         & 70.5                         \\
                                                                                                     & AMD \cite{10191080}                          & 42.4 (+4.0)                         & 24.1                         & 46.5                         & 56.2                         & 55.8 (+3.8)                         & 35.3                         & 60.0                         & 70.8                         \\
\multirow{-6}{*}{\begin{tabular}[c]{@{}c@{}}Cascade\\ Mask RCNN~\cite{cai2019cascade}\\ ResNeXt101\\ (47.3)\end{tabular}} & \cellcolor[HTML]{EFEFEF}\textbf{DMKD} (Ours) & \cellcolor[HTML]{EFEFEF}\textbf{42.7} (\textcolor{blue}{+4.3}) & \cellcolor[HTML]{EFEFEF}24.4 & \cellcolor[HTML]{EFEFEF}46.9 & \cellcolor[HTML]{EFEFEF}56.5 & \cellcolor[HTML]{EFEFEF}\textbf{56.0} (\textcolor{blue}{+4.0}) & \cellcolor[HTML]{EFEFEF}35.5 & \cellcolor[HTML]{EFEFEF}60.4 & \cellcolor[HTML]{EFEFEF}71.0 \\ \hline
\end{tabular}
\label{results}
}
\end{table*}

\subsection{Preliminaries}
Before elaborating on our proposed method, we briefly review the basic feature-based distillation framework formulated as:
\begin{equation}
\mathcal{L}_{base}=\sum_{c=1}^{C}\sum_{h=1}^{H}\sum_{w=1}^{W}\textit{MSE}\left(
\Theta_{align}(\mathrm{F}_{c;\,h;\,w}^{S}),\mathrm{F}_{c;\,h;\,w}^{T}
    \right)
\end{equation}
where $\mathrm{F}_{c;\,h;\,w}^{S/T}\in \mathbb{R}^{C\times H\times W}$ is the feature map generated from the student or teacher network. $MSE(\cdot)$ denotes the fitting error between the student feature aligned by a learnable function $\Theta_{align}$ and the teacher feature. It can be observed from the above formulation that the vanilla feature-based distillation model forces the student raw feature to be as close as possible to the teacher counterpart. In this sense, the teacher acts as a template to be mimicked directly and feature distillation can be identified as a straightforward and no-frills algorithm because the features of student and teacher networks are directly used for representational matching.

\subsection{Our DMKD framework}

To comprehensively encode object-aware semantics into the student network, we develop a dual masked knowledge distillation (DMKD) framework in which both the spatially salient regions and informative channels are uncovered for preferable masked feature reconstruction. Notably, our model can be treated as a plug-and-play augmentation module that is applicable to the vanilla feature-based distillation methods. As shown in Fig.~\ref{dmkd_arch}, our DMKD consists of three key steps: dual attention map generation, attention-guided masking, and self-adjustable weighted fusion. To begin with, the spatial and channel-wise attention maps are produced from the teacher feature formulated as follows:
\begin{align}
    &A^{s}=\phi_{align}\left(Sigmoid\left(
    \frac{1}{C\mathcal{T}}\left\langle \left\|\mathrm{F}^{T}_{1}\right \|_{2}^{2}, ..., \left\|\mathrm{F}^{T}_{n}\right \|_{2}^{2}\right\rangle
    \right)\right) \label{as} \\
    &A^{c}=Sigmoid\left(
    \frac{1}{HW\mathcal{T}}\sum_{h=1}^{H}\sum_{w=1}^{W}\left\langle\mathrm{F}_{h,w,1}^{T},..., \mathrm{F}_{h,w,C}^{T}\right\rangle
    \right) \label{ac}
\end{align}
where $A^{s}\in \mathbb{R}^{1\times H\times W}$ and $A^{c}\in \mathbb{R}^{C\times 1\times 1}$ are spatial and channel-wise attention map, respectively. $\mathrm{F}^{T}_{n}$ in Eq.~(\ref{as}) is the $n$-th vector of teacher's feature map, while $\mathrm{F}^{T}_{h,w,C}$ in Eq.~(\ref{ac}) is the feature map of $C$-th channel. $\mathcal{T}$ is introduced for adjusting the distribution~\cite{hinton2015distilling}.

Next, the resulting dual attention maps $A^{s}$ and $A^{c}$ are used for deriving the corresponding masking maps $M^{s}$ and $M^{c}$ with the thresholds $\tau_{s}$ and $\tau_{c}$:
\begin{align}
    M^{s}_{i\subseteq [0, H), j\subseteq [0, W)}&=
    \left\{\begin{aligned}
    & 0, \quad A_{i,j}^{s}\geq \tau_{s} \\
    & 1, \quad Otherwise
    \end{aligned}\right. \\
    M^{c}_{k\subseteq [0, C)}&=\left\{\begin{aligned}
    & 0, \quad A_{k}^{c}\geq \tau_{c} \\
    & 1, \quad Otherwise
    \end{aligned}\right.
\end{align}
where $M^{s}$ and $M^{c}$ share the same shape with $A^{s}$ and $A^{c}$, respectively. Then, we utilize them to perform spatial and channel masking on the student feature $\mathrm{F}^{S}$, generating two masked features
$\mathrm{F}^{S}_{s}$ and $\mathrm{F}^{S}_{c}$:
\begin{align}
& \mathrm{F}^{S}_s=\Theta_{align}\left(\mathrm{F}^{S}\right) \otimes M^{s} \\
& \mathrm{F}^{S}_c=\Theta_{align}\left(\mathrm{F}^{S}\right) \otimes M^{c}
\end{align}
where $\otimes$ denotes the element-wise multiplication.

Finally, two different generation blocks based on convolution and Multilayer Perceptrons (MLPs) are respectively employed to reconstruct the spatial and channel-wise masked features. Notably, the channel reconstruction is achieved by MLPs, which differs from the spatial one as in MGD and AMD because the channel features are spatially independent. In other words, this generative layer should only focus on the interaction of inter-channels which will be demonstrated in the ablation study. In addition, pairwise self-adjustable weighting factors $\alpha$ and $\beta$ are introduced to fuse the dual reconstructed features, leading to the final student feature formulated as:
\begin{equation} \label{rec}
    \mathrm{F}^{rec}=\alpha\cdot \Theta_{Conv}(\mathrm{F}^{S}_{s}) + \beta\cdot \Theta_{MLP}(\mathrm{F}^{S}_{c})
\end{equation}
Thus, the distillation loss can be built on the reconstructed student feature $F^{rec}$ and the teacher counterpart $F^T$ as follows: 
\begin{equation} \label{dmkd_loss}
\mathcal{L}_{DMKD}=\sum_{c=1}^{C}\sum_{h=1}^{H}\sum_{w=1}^{W}\textit{MSE}\left(
\mathrm{F}_{c;\,h;\,w}^{rec},\mathrm{F}_{c;\,h;\,w}^{T}
    \right)
\end{equation}
\section{Experiments}
\subsection{Experimental Setup}

We have evaluated our DMKD on the public MS COCO2017 \cite{lin2014microsoft} benchmarking dataset for the object detection task. Besides, Average Precision (AP) and Average Recall (AR) are used for performance measures. 
The hyperparameters are empirically set to $\tau_{s}=0.55$, $\tau_{c}=0.65$, $\alpha=\beta=0.5 (initial)$, $\mathcal{T}=0.5$. Besides, $\gamma$ is set as $5\times 10^{-7}$ and $5\times 10^{-6}$ for the one-stage and the two-stage models respectively. During the training process, we employ the SGD optimizer to train all detectors over the course of 24 epochs. Besides, ResNeXt101~\cite{xie2017aggregated} and ResNet50~\cite{he2016deep} are used as the respective backbones for the teacher and the student networks.

\subsection{Results}

As shown in Table \ref{results}, in the first group of experiments, the results indicate that our DMKD not only significantly boosts the baseline student model by 4.1\% mAP but also outperforms the teacher network by 0.5\% mAP. The advantages of DMKD against the state-of-the-art MGD and AMD are manifested by respective further performance gains of 0.5\% and 0.2\%. Similar performance superiority of our method can also be observed regarding mAR metric. When the RetinaNet network is replaced with the other detectors in the remaining experiments, analogous results are also reported.

\subsection{Ablation studies}
In this section, we conduct different ablation studies to gain further insight into the interior design of our DMKD framework with RetinaNet used as the detector.
\begin{table}[tb]
\centering
\caption{Ablation study of different \textbf{masking strategies}. $C$ and $S$ denote single Channel and Spatial-wise masking respectively.} 
\resizebox{0.85\columnwidth}{!}{%
\begin{tabular}{ccccc} \toprule[0.5pt]
\begin{tabular}[c]{l} \textit{Masking strategy}\end{tabular} & \textit{Dual (Ours)} & \textit{w/o C} & \textit{w/o S} & \textit{w/o masking} \\ \midrule[0.5pt]
mAP (\%)                                                                       & 41.5        & 41.3 \textcolor{blue}{(-0.2)}  & 41.2 \textcolor{blue}{(-0.3)}  & 39.6 \textcolor{blue}{(-1.9)}     \\ \bottomrule[0.5pt]
\end{tabular}
}
\label{ab_1}
\end{table}
\subsubsection{Masking strategy}    
We first explore the effect of feature masking on the distillation performance and compare different masking strategies. As shown in Table~\ref{ab_1}, a significant performance drop of exceeding 1.5\% can be observed when any masking strategy is not involved, suggesting that masking strategy plays a beneficial role in feature distillation for enhanced feature reconstruction. Furthermore, when comparing our dual masking scheme with both of the single masking strategies, consistent performance gains are reported, demonstrating the advantage of dual masking in capturing both spatially salient regions and informative channel-wise clues for encoding comprehensive object-aware representations.
\begin{table}[tb]
\centering
\caption{Ablation study of \textbf{spatial and channel masking thresholds} $\tau_{s}$ and $\tau_{c}$.}
\resizebox{0.9\columnwidth}{!}{%
\begin{tabular}{cccc}\toprule[0.5pt]
 $\tau_{s}$ ($\tau_{c}=0.65$) & 0.55 & 0.45 (mask ratio $\uparrow$) & 0.65 (mask ratio $\downarrow$) \\ \midrule[0.5pt]
mAP (\%)         & 41.5    & 41.5 (\textcolor{blue}{+0.0})    & 41.4 (\textcolor{blue}{-0.1})    \\ \midrule[0.5pt]
$\tau_{c}$ ($\tau_{s}=0.55$) & 0.65 & 0.55 (mask ratio $\uparrow$)& 0.75 (mask ratio $\downarrow$) \\ \midrule[0.5pt]
mAP (\%)         & 41.5    & 41.3 (\textcolor{blue}{-0.2})    & 41.4 (\textcolor{blue}{-0.1})    \\ \bottomrule[0.5pt]
\end{tabular}
\label{ab_2}
}
\end{table}
\subsubsection{Masking ratios}
Table~\ref{ab_2} illustrates the performance of our method with varying masking ratios. It can be clearly observed that the model performance slightly deteriorates when decreasing the spatially masking ratio with increased $\tau_s$, which is consistent with the finding in MAE~\cite{he2022masked}. This implies some spatially important regions may be excluded from spatial masking. On the other hand, when increasing the channel-wise masking ratio, a similar performance decline can also be observed, suggesting that the model excessively focuses on redundant and non-informative channels. In our experiments, the dual thresholds $\tau_s$ and $\tau_c$ are empirically set as 0.55 and 0.65.


\subsubsection{Feature generative blocks}
\begin{table}[tb]
\centering
\caption{Ablation study of two \textbf{different generative blocks} for channel reconstruction.}
\resizebox{0.85\columnwidth}{!}{%
\begin{tabular}{ccc}\toprule[0.5pt]
\textit{Generative block} & \textit{Conv-based} $\Theta_{Conv}$ & \textit{MLP-based} $\Theta_{MLP}$ \\ \midrule[0.5pt]
mAP (\%)                & 41.0 (\textcolor{blue}{-0.5})           & 41.5 \\         \bottomrule[0.5pt]
\end{tabular}%
}
\label{ab_3}
\end{table}
We also discuss the effect of feature generation block design on channel-wise feature reconstruction. As shown in Table 4, the MLP-based module within our DMKD achieves better performance than the Conv-based one. This can be explained by the properties of the MLP operator, \textit{i.e.,} in contrast with the spatial interaction as in ConvNets, it acts independently on the spatial dimension and only models the inter-channel relations. Specifically, we use two successive linear projections $\theta_{MLP}^{'}\in \mathbb{R}^{C\times 2C}$ and $\theta_{MLP}^{''}\in \mathbb{R}^{2C\times C}$ followed by GELU~\cite{hendrycks2016gaussian} and Layer Normalization~\cite{ba2016layer} in practice.

\section{Conclusion}
In this paper, we explore the masked-based distillation algorithms and propose a simple and effective feature augmentation framework termed dual masking knowledge distillation (DMKD). In contrast with previous masked-based algorithms, we simultaneously focus on spatial and channel dimensions which respectively reflect the important spatial regions and object-aware information. Moreover, DMKD is a dynamic algorithm that adaptively captures the most semantically relevant spatial areas and informative channels, improving flexibility over the previous methods. Extensive experiments demonstrate the superiority of our design.


\bibliographystyle{IEEEbib}
\bibliography{refs}

\end{document}